\documentclass[a4paper, 10pt, conference]{ieeeconf}      
\usepackage{FG2021}
\usepackage{multirow}

\usepackage[accsupp]{axessibility}

\usepackage{graphicx}
\usepackage{subfig}

\usepackage{float}
\usepackage{mathtools, nccmath}
\usepackage{amsmath,amssymb}
\DeclareMathOperator{\E}{\mathbb{E}}
\FGfinalcopy 

\newcommand{\changetext}[1]{\textcolor{black}{#1}}

\IEEEoverridecommandlockouts                              
\def\FGPaperID{124} 

\title{\LARGE \bf
The Imaginative Generative Adversarial Network: Automatic Data Augmentation for Dynamic Skeleton-Based Hand Gesture and Human Action Recognition}

\author{\parbox{16cm}{\centering
    {\large Junxiao Shen, John Dudley and Per Ola Kristensson}\\
    {\normalsize
    Department of Engineering, University of Cambridge, United Kingdom}}
    \thanks{John Dudley and Per Ola Kristensson were supported by EPSRC (grant EP/S027432/1).}
}

\begin{document}

\IEEEoverridecommandlockouts\pubid{\makebox[\columnwidth]{978-1-6654-3176-7/21/\$31.00~\copyright{}2021 IEEE \hfill}
\hspace{\columnsep}\makebox[\columnwidth]{ }}

\ifFGfinal
\thispagestyle{empty}
\pagestyle{empty}
\else
\author{Anonymous FG2021 submission\\ Paper ID \FGPaperID \\}
\pagestyle{plain}
\fi
\maketitle


\begin{abstract}
    Deep learning approaches deliver state-of-the-art performance in recognition of spatiotemporal human motion data. However, one of the main challenges in these recognition tasks is limited available training data. Insufficient training data results in over-fitting and data augmentation is one approach to address this challenge.
    Existing data augmentation strategies based on scaling, shifting and interpolating offer limited generalizability and typically require detailed inspection of the dataset as well as hundreds of GPU hours for hyperparameter optimization.
    In this paper, we present a novel automatic data augmentation model, the Imaginative Generative Adversarial Network (GAN), that approximates the distribution of the input data and samples new data from this distribution. It is automatic in that it requires no data inspection and little hyperparameter tuning and therefore it is a low-cost and low-effort approach to generate synthetic data.
    We demonstrate our approach on small-scale skeleton-based datasets with a comprehensive experimental analysis. Our results show that the augmentation strategy is fast to train and can improve classification accuracy for both conventional neural networks and state-of-the-art methods.
\end{abstract}

\section{INTRODUCTION}

\begin{table*}[t]
\centering
\begin{tabular}{|c|c|c|c|c|c|c|c|}
\hline
\textbf{Dataset}                            & \textbf{Model}                    & \textbf{Augmentation} & \textbf{Accuracy} & \textbf{Improvement} & \textbf{Standard Error} & \textbf{Time (hrs)} & \textbf{Contribution}                                                                                                \\ \hline
\multirow{6}{*}{SHREC'17 Track} & \multirow{3}{*}{LSTM} & CD  & $72.0\%$ & $\cdot$          & $\pm0.94\%$   & $\cdot$   & \multirow{3}{*}{\begin{tabular}[c]{@{}c@{}}15 times faster,\\ Accuracy increased by $3.5\%$\end{tabular}}  \\ \cline{3-7}
                                &                          & CAD & $76.6\%$ & $4.6\%$     & $\pm 2.04\%$   & 75.2 &                                                                                                             \\ \cline{3-7}
                                &                          & GAD & $80.1\%$ & $8.1\%$     & $\pm 0.64\%$   & 5    &                                                                                                             \\ \cline{2-8} 
                                & \multirow{3}{*}{CNN}     & CD  & $4.30\%$ & $\cdot$          & $\pm 0.00 \%$  & $\cdot$   & \multirow{3}{*}{\begin{tabular}[c]{@{}c@{}}5 times faster,\\ Accuracy increased by $1.8\%$\end{tabular}}   \\ \cline{3-7}
                                &                          & CAD & $78.0\%$ & $73.7\%$    & $\pm 0.77 \%$  & 25.2 &                                                                                                             \\ \cline{3-7}
                                &                          & GAD & $79.8\%$ & $75.5\%$    & $\pm 0.57\%$   & 5    &                                                                                                             \\ \hline
\multirow{6}{*}{MSR Action3D}  & \multirow{3}{*}{LSTM} & CD  & $18.4\%$ & $\cdot$          & $\pm 10.64 \%$ & $\cdot$   & \multirow{3}{*}{\begin{tabular}[c]{@{}c@{}}5 times faster,\\ Accuracy increased by $8.8\%$\end{tabular}}  \\ \cline{3-7}
                                &                          & CAD & $58.5\%$ & $40.1\%$    & $\pm 6.54\%$   & 11.2 &                                                                                                             \\ \cline{3-7}
                                &                          & GAD & $67.3\%$ & $48.9\%$    & $\pm 0.85 \%$   & 2.1  &                                                                                                             \\ \cline{2-8} 
                                & \multirow{3}{*}{CNN}     & CD  & $4.73\%$ & $\cdot$          & $\pm 0.42\%$   & $\cdot$   & \multirow{3}{*}{\begin{tabular}[c]{@{}c@{}}2 times faster,\\ Accuracy increased by $54.04\%$\end{tabular}} \\ \cline{3-7}
                                &                          & CAD & $5.76\%$ & $1.03\%$       & $\pm 1.12\%$   & 4.1 &                                                                                                             \\ \cline{3-7}
                                &                          & GAD & $59.8\%$ & $55.07\%$   & $\pm 2.62\%$   & 2.1  &                                                                                                             \\ \hline
\end{tabular}

\caption{The \textbf{Augmentation} column describes three different types of data: clean data (CD), which is denoised and padded raw data; classical augmented data (CAD), which is data augmented using a classical approach; and GAN-augmented data (GAD), which is data generated from the Imaginative GAN. The data is used to train two different models, the first is LSTM-based and the second is CNN-based. The data is from two public datasets, the SHREC'17 Track dataset and the MSR Action3D dataset. \textbf{Accuracy} is the mean validation accuracy of the trained model performed on CD, CAD, or GAD dataset. \textbf{Improvement} is the absolute improvement in accuracy resulting from the augmented data (either CAD or GAD) compared to the clean data (CD). \textbf{Standard Error} is the standard deviation of validation accuracy divided by the square root of the number of seeds (we use four random seeds). \textbf{Time} is the time taken for the hyperparameter optimization of the classical approach, or the time taken for the Imaginative GAN to converge. \textbf{Contribution} is the improvement brought by GAD for both accuracy and time compared to CAD. The data augmented by the Imaginative GAN provides higher accuracy and is more time-efficient.} 
\label{table:1}
\end{table*}

Hand gesture interaction has the potential to provide users with a fluid and unencumbered method of interfacing with computer systems. As a consequence, it has attracted considerable research attention in both hand gesture recognition (HGR) and deployment of such strategies to diverse domains including smart homes, augmented reality, virtual reality, manufacturing, and smart cars. 
Human action recognition (HAR) is another popular research area with many real-world applications, such as surveillance event detection, video retrieval, and smart rehabilitation~\cite{poppe2010survey}.
One of the main challenges in HGR/HAR research is that factors, such as the complexity of hand gesture structures/action structures, differences in hand size/human size, and hand postures/human postures, can influence the performance of the recognition algorithm. While deep neural networks have had remarkable success in HGR~\cite{chen2019construct,yang2019make,yang2019make} and  HAR~\cite{ullah2017action,ding2017investigation}, problems in over-fitting or a failure to learn a high-performance model may arise when training deep neural networks with insufficient training data. \changetext{
However, increasing the amount of training data through data augmentation can help alleviate this issue and improve model performance.
}


Classical data augmentation methods increase the volume of the training set by applying transformations to raw data, which may be realistic or unrealistic, such as SpecAugment~\cite{park2019specaugment}, cutout~\cite{devries2017improved} and mixup~\cite{zhang2017mixup}. However, these methods are not always effective on all datasets. 
Further, the above classical data augmentation methods are limited because they are structured such that they transform existing samples into slightly altered additional samples. Moreover, hyperparameter optimization for a classical augmentation policy usually requires a large computation budget using a brute force approach that can easily cost hundreds of GPU hours. In addition, there are many different ways of altering the samples. This highlights a need to find a way of generating new data efficiently.

\changetext{
Another important requirement for data augmentation methods is that they produce diversity in data which is representative of natural user variation.
Achieving this through simple transformations, such as rotation, scaling and noise injection, is non-trivial.
While such methods act as a perturbation model that increase a neural network's robustness, this may not necessarily translate into better generalization.
Ideally, the distribution of intra and inter user variability is captured and reflected by the data augmentation method.}
In this paper, we propose a high-efficiency sampling strategy that can directly estimate the training data distribution and generate new samples based on the estimated distribution.




Generative Adversarial Networks (GANs) have gained much popularity in modeling data distributions directly.
A GAN is a powerful tool to generate unobserved data using a minimax game without supervision~\cite{goodfellow2014generative}. Inspired by the recent success of GAN-based data augmentation in speech and vision domains~\cite{frid2018gan,hu2018generative}, we propose a novel variant we call the \emph{Imaginative GAN} which assists in discovering an approximation of the true distribution of input data. The novelty of this work is the following:
\begin{enumerate}
    \item We propose a data-efficient unsupervised learning model for skeleton-based data augmentation: the Imaginative GAN. A trained Imaginative GAN can generalize to new data with unseen classes. 
    
    \item Since it does not require prior knowledge and inspection of input data for training, the Imaginative GAN allows an automatic and cross-domain data augmentation process with little hyperparameter tuning.
\end{enumerate}


\changetext{The Imaginative GAN leverages a CycleGAN~\cite{zhu2017unpaired} structure and extends prior work with two novel modifications.
First, CycleGAN was originally proposed to transfer latent attributes between two domains whereas we use it here to add variability to data within the same domain.
Second, we introduce an unsupervised training strategy reminiscent of teacher forcing that incorporates the ground truth input data at each time step.
These novel modifications improve training stability and permit using roughly one tenth of the amount of training data normally required.}

We evaluate our approach on two public datasets (the SHREC'17 Track dataset~\cite{SHREC2017De} and the MSR Action3D dataset~\cite{li2010action}) using two recognition models: an LSTM and a CNN modified from prior work~\cite{nunez2018convolutional,lai2018cnn}. We compare performance resulting from using augmented data from the Imaginative GAN (GAD) with classically augmented data (CAD). The classical approach adds realistic transformations to the data, including scaling, shifting, interpolating, and adding Gaussian noise. This approach is described in detail in Section \ref{Generating}. 
As a reference, we also evaluate the effects of using the denoised and padded raw data without any augmentation, a condition we refer to as clean data (CD). 
We evaluate performance using two main criteria. 
First, the time required to find the optimized strategy for recognition accuracy in the classical approach using grid search, or the time required for the Imaginative GAN to converge. Second, the validation accuracy on each type of data.

In summary, we will demonstrate that the Imaginative GAN gives rise to the following four key properties:

\begin{enumerate}
    \item \textbf{Higher accuracy}: As shown in Table \ref{table:1}, the recognition models trained on GAN-augmented data (GAD) achieve the best validation accuracy for both the SHREC'17 Track and MSR Action3D datasets.
    \item \textbf{Increased stability}: Performance is more stable on models trained on GAD compared to models trained on CD and CAD, which is observed in the smaller standard error (see Table \ref{table:1}). 
    \item \textbf{Temporal efficiency}: Table \ref{table:1} shows that the time taken for the classical approach to find the best combination of hyperparameters is long and varies depending on the choice of recognition model. If the model is large and difficult to converge, the time required is increased. In contrast, the Imaginative GAN is fast to train and the augmentation strategy is decoupled from the recognition model. It is possible to reduce time in the classical approach using a coarser grid search. However, as a result, the validation accuracy may then decrease due to suboptimal hyperparameters. Therefore, in the classical approach there is a time-accuracy trade-off. 
    \item \textbf{Generalization to new classes}: The Imaginative GAN is effective in generalizing to data with unseen classes. In other words, as long as the data to train the Imaginative GAN is in the same domain as the data with new classes, the trained model is able to generate realistic samples of the new data.  
\end{enumerate}



\section{Related Work}
\label{relatedwork}
The term \emph{data augmentation} originated from Tanner and Wong~\cite{Martin1987Calculation}, linking augmented data with observed data via a many-to-one mapping $\textsl{M}: Y_{aug}\rightarrow Y_{obs}$. There are different types of data augmentation strategies. A simple but effective approach is to add Gaussian noise, which is not particularly domain-specific yet can help prevent the model from over-fitting~\cite{lopes2019improving}. Other techniques use various realistic transformations, such as elastic distortions, as well as distortions in scale, orientation, and position of training images~\cite{cubuk2018autoaugment}. Color adjustment, blurring and sharpening, white balance, and other distortions have also been previously used on CIFAR-10 and ImageNet~\cite{krizhevsky2012imagenet, zoph2018learning}. Unrealistic distortions, such as cutout~\cite{devries2017improved} and mixup~\cite{zhang2017mixup}, help regularize the training of the neural networks. However, these classical augmentation approaches usually require prior knowledge of the data and inspection of the data in addition to time-consuming hyperparameter optimization.  

An alternative strategy is to augment the training set by modeling the data distribution directly using generative models, such as Bayesian approaches or GANs~\cite{tran2017bayesian}. GANs can learn generative models via adversarial training to produce samples from the approximated real data distribution. Different variants of GANs have been proposed to generate realistic natural images~\cite{xu2018attngan,frid2018synthetic}.
In speech processing, GANs have been applied to speech synthesis~\cite{kaneko2017generative}, acoustic scene classification~\cite{mun2017generative}, and speech recognition~\cite{hu2018generative}. 
GANs have also been used to generate synthetic path data to improve the performance of a continuous path keyboard~\cite{mehra2020leveraging} \changetext{and to jointly optimize data augmentation and network training in human pose estimation from images~\cite{peng2018jointly}.}
To our knowledge, there is no prior work investigating using GANs to synthesize skeleton data for data augmentation.



\section{Generating Synthetic Human Motion Data}
\label{Generating}
In this paper, we compare two augmentation strategies. The first is a classical data augmentation strategy, which uses different sequential and stochastic transformations. The transformations are sometimes arbitrary and arise from manual inspection of the data. Usually, these strategies cannot be transferred across different data domains. The second strategy is the Imaginative GAN, which approximates the real data distribution and then samples from this approximated distribution. 

\subsection{Classical Data Augmentation}
From inspection of the raw skeleton data, we observe natural variation in aspects such as size, posture and the speed of performing the actions/gestures. Thus such variations can be applied as transformations. We modified the data augmentation approach from Núñez et al.~\cite{nunez2018convolutional} to make transformations more realistic. 
The major difference is that we sample the factors from a Gaussian distribution instead of a Uniform distribution. This is because our preliminary experiments revealed that a Gaussian distribution leads to better validation recognition accuracy with lower standard error than a Uniform distribution. 
Acknowledging that there are hundreds of specifications a researcher may choose to use, we have chosen the following specifications to simulate the process of how a researcher would approach the data augmentation problem using a classical approach, balancing subjective realism in transformations and a desire for high objective performance: 
\begin{enumerate}
    \item \textbf{Scale}: A scale factor, sampled from $\mathcal{N}(1, \, \sigma^{2}_{\text{scale}})$, is applied to each data point globally, where $\sigma_{\text{scale}}$ is a hyperparameter.
    \item \textbf{Shift}: A global displacement $(d_{x},d_{y},d_{z})$, sampled from $\mathcal{N}(0, \, \sigma^{2}_{\text{shift}})$, is applied to all the data points, where $\sigma_{\text{shift}}$ is a hyperparameter. 
    \item \textbf{Time Interpolation}: Cubic interpolation is used to interpolate the skeleton sequence. The interpolated positions are sampled from a Uniform distribution over the length of the sequence.  
    \item \textbf{Noise}: A number of joints are randomly selected and then Gaussian noise sampled from $\mathcal{N}(0,\sigma^{2}_{\text{noise}})$ is applied to the positions, where $\sigma_{\text{noise}}$ is a hyperparameter. The number of joints selected is randomly generated between 1 to 8 for the SHREC'17 Track dataset and 1 to 4 for the MSR Action3D dataset. The number here is chosen by preliminary experiments to decrease the total number of hyperparameters. The noise added to each selected joint is different. However, the noise is the same at every timestep along the entire sequence.
\end{enumerate}

\begin{figure}[t]
  \centering
  \subfloat[The Imaginative GAN is trained using a CycleGAN structure with a cycle consistency loss. The latent attributes are transferred within the input datasets $X$ and $Y$. $G$ and $F$ are the two generators, $D_{X}$ and $D_{Y}$ are the two discriminators. ]{\includegraphics[width=0.23\textwidth]{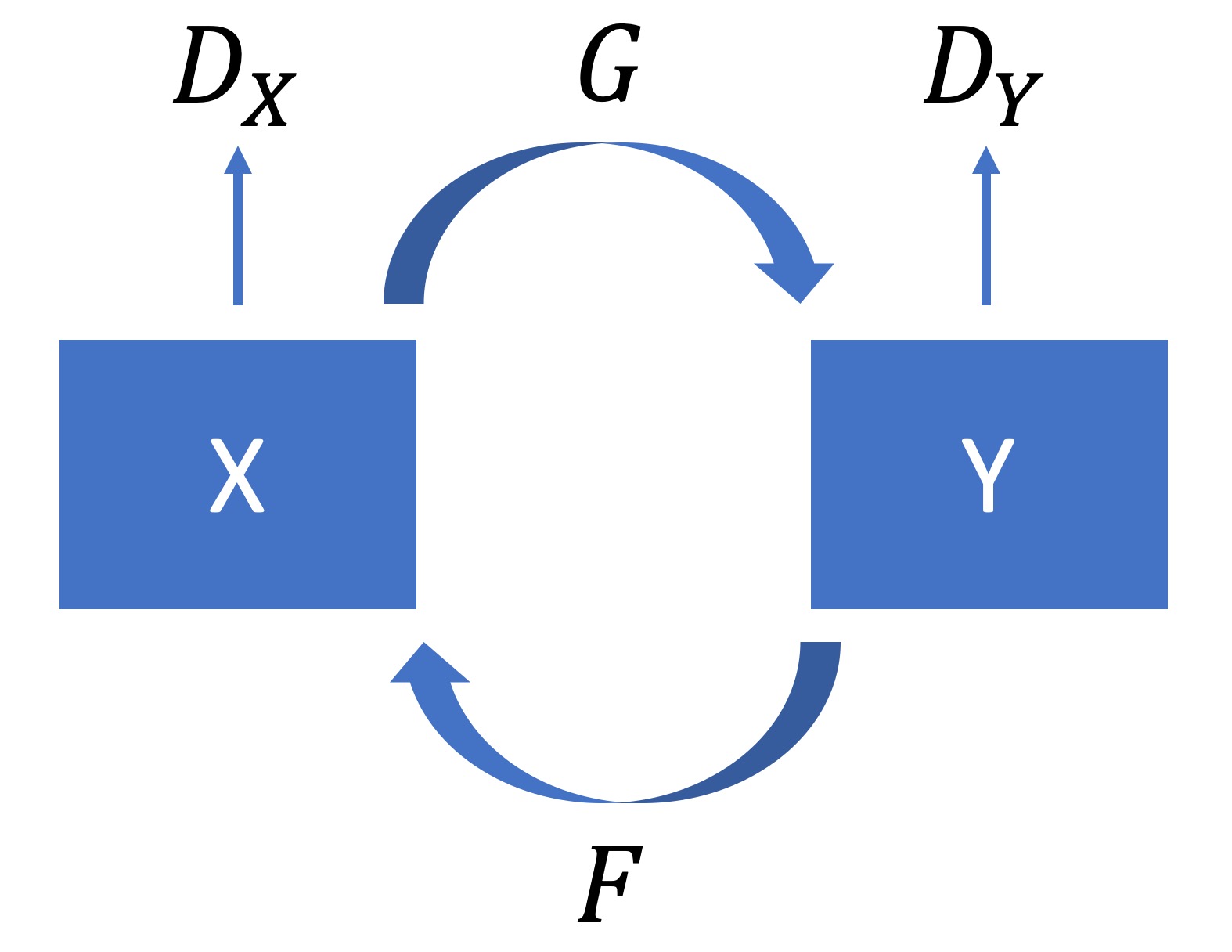}\label{fig:cyclegan}}
  \hfill
  \subfloat[The generator in the Imaginative GAN has one GRU layer and one fully connected layer. It has a teacher forcing-like mechanism so the input is a ground truth sample and the output is an altered sample. $T$ is the length of the sample.]{\includegraphics[width=0.23\textwidth]{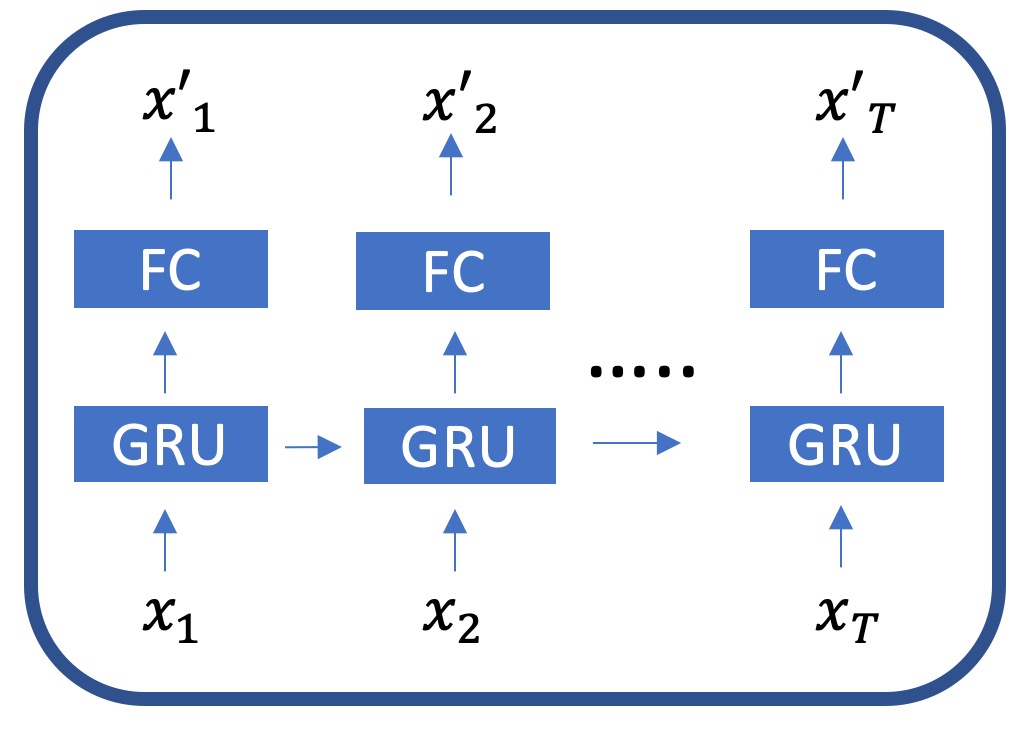}\label{fig:generator}}
  \caption{Structure of the Imaginative GAN and its generator.}
\end{figure}

\subsection{Imaginative GAN}
A GAN can be characterized as training a pair of networks competing against each other---it is a minimax game~\cite{goodfellow2014generative}. In a GAN, a \emph{generator} attempts to generate as real samples as possible while a \emph{discriminator} attempts to distinguish between the generated and real samples. Conventional GAN structures require noise as input. Training a conventional GAN is challenging because not only do the model parameters oscillate, destabilize, and sometimes never converge, but in addition the generator may collapse and produce a limited variety of samples. A large amount of data and time-consuming hyperparameter selections are required to train effective GAN models. To tackle the above challenges, we introduce a mechanism reminiscent of teacher forcing in the generator.
\changetext{This mechanism uses ground truth data as the input at each timestep. 
This leads to the Imaginative GAN converging more easily and demanding less data overall.}
\changetext{The distinction from typical teacher forcing is that we are not seeking to force the output towards some target, but rather to facilitate augmentations applied with respect to the ground truth sample.}
We also utilize the CycleGAN structure in transferring the latent attributes within the dataset, unlike the original work for unpaired image translation~\cite{zhu2017unpaired}, which transfers styles between two different datasets.


\subsubsection{Networks}
The Imaginative GAN has a similar structure to CycleGAN, as shown in Figure \ref{fig:cyclegan}. It has two GAN structures: each GAN has one generator and one discriminator. The generator is shown in Figure \ref{fig:generator}. The generator has a Gated Recurrent Unit (GRU) layer and a fully connected layer. No activation function is used in the fully connected layer. A GRU makes each recurrent unit adaptively capture dependencies of different time scales~\cite{chung2014empirical}. Unlike an LSTM unit, a GRU does not have separate memory cells. The generator has a mechanism reminiscent of teacher forcing that uses the ground truth data as input for every GRU cell. Teacher forcing is often used in the training of sequence to sequence models, such as machine translation models. To our knowledge, this is the first time that a teacher forcing mechanism is used in this way in a GAN. The use of teacher forcing enables the model to easily converge and become fast to train. The input sequence to the generator is $[x_{1},x_{2},...,x_{T}]$ where $T$ is the length of the sequence and the output sequence of the generator is $[x'_{1},x'_{2},...,x'_{T}]$. The discriminator has one fully connected layer 
attached to a network identical to the generator to output a scalar for discrimination. 


\subsubsection{Formulation}
The goal of the Imaginative GAN is to learn the latent attributes, such as behavioral attributes (speed of performing the actions/gestures etc.,) and physical attributes (human/hand sizes etc.,). Thereafter, these learned latent attributes are applied to other data. That is, in mathematical terms, we have two generators to learn two mapping functions, $G:X\rightarrow Y$ and $F:Y\rightarrow X$, between two domains, $X$ and $Y$, given samples $\left \{x_{i}\right \}^{N}_{i=1}$ where $x_{i} \in X$ and $\left \{y_{j}\right \}^{N}_{j=1}$ where $y_{j} \in Y$. 
In our case, each domain represents a separate partition of the dataset.
The model also has two discriminators $D_{X}$ and $D_{Y}$ to discriminate between the generated sample ${F(y)}$ or ${G(x)}$ and the real data ${x}$ or ${y}$. The generator $G$ tries to generate synthetic data $G(x)$ that has similar attributes from domain $Y$ with objective:
\begin{align}
\begin{split}\label{eq:gan_gen}
        \mathcal{L}_{gen}(G, D_{Y}, X) ={}& \E_{x\sim p_{data}(x)}\left [\log D_{Y}(G(x)) \right ]
\end{split}
\end{align}

The discriminator $D_{Y}$ tries to distinguish $G(x)$ from real data $y$ with objective:


\begin{align}
\begin{split}\label{eq:gan_dis}
        \mathcal{L}(G, D_{Y}, X, Y) ={}& \E_{y\sim p_{data}(y)}\left [\log D_{Y}(y) \right ]\\
         & +\E_{x\sim p_{data}(x)}\left [\log (1-D_{Y}(G(x))) \right ]
\end{split}
\end{align}

Cycle consistency loss is used to encourage $F(G(x))\approx x$ and $G(F(y))\approx y$~\cite{zhu2017unpaired}:

\begin{align}
\begin{split}\label{eq:cyc}
        \mathcal{L}_{cyc}(G, F) ={}& \E_{y\sim p_{data}(y)}\left [ \left \| G(F(y))-y \right \|_{1} \right ]\\
         & +\E_{x\sim p_{data}(x)}\left [ \left \| F(G(x))-x \right \|_{1} \right ]
\end{split}
\end{align}

Noise is injected at each translation step and identity loss is introduced to reduce the noise: 
\begin{align}
\begin{split}\label{eq:iden}
        \mathcal{L}_{identity}(G) ={}& \E_{y\sim p_{data}(y)}\left [ \left \| G(y)-y \right \|_{1} \right ]\\
         & +\E_{x\sim p_{data}(x)}\left [ \left \| G(x)-x \right \|_{1} \right ]
\end{split}
\end{align}

The full objective for the generator $G$ is:

\begin{align}
\begin{split}\label{eq:all}
        \mathcal{L}(G, F, D_{Y}) ={}& \mathcal{L}_{gen}(G,D_{Y},X)\\
      & + \lambda_{1}\mathcal{L}_{cyc}(G,F)\\
      & + \lambda_{2}\mathcal{L}_{identity}(G)
\end{split}
\end{align}
where $\lambda_{1}$ and $\lambda_{2}$ are the two weights of the losses. 
We are using different partitions of the same dataset for $\left \{x_{i}\right \}^{N}_{i=1}$ and $\left \{y_{i}\right \}^{N}_{i=1}$.
The generator is then effectively approximating the input data distribution through optimization by stochastic gradient descent in mini-batches.





\section{Experiments}


\changetext{Affinity and diversity are two interpretable, easy-to-compute metrics used for parameterizing augmentation performance~\cite{gontijo2020affinity} and 
to quantitatively evaluate
the properties of the three types of data: clean data (CD), classical augmented data (CAD), and GAN augmented data (GAD).}
\begin{enumerate}
    \item \textbf{Affinity}: Affinity quantifies the shift between the clean data distribution and the augmented data distribution. Affinity is calculated using the definition from Gontijo-Lopes et al.~\cite{gontijo2020affinity}: the difference between the validation accuracy of a model trained on clean data and tested on clean data and the validation accuracy of the same model tested on an augmented validation set. 
    \item \textbf{Diversity}: Diversity quantifies the complexity of the augmented data with respect to the model and optimization procedure~\cite{gontijo2020affinity}. In this paper, diversity is calculated as the difference between the validation loss and the training loss. The intuition is that the more diverse the data is, the higher the difference between the training loss and validation loss since the distance between the training data and validation data distribution will be larger. This is different from the method in Gontijo-Lopes et al.~~\cite{gontijo2020affinity}, where they solely use the training loss. However, only using training loss is model dependent, and may also be data dependent in certain circumstances. Therefore, to compare the two very different data augmentation polices, we use the difference between training loss and validation loss since it is more consistent and independent. 
\end{enumerate}

We would like the augmented data to have a high affinity and a reasonable level of diversity. The reasoning of how affinity and diversity affect the relationship between the augmented data and clean data is clearly described and illustrated by Gontijo-Lopes et al.~\cite{gontijo2020affinity}.

\subsection{Evaluation Datasets}
There are several public dynamic gesture datasets~\cite{marin2014hand,SHREC2017De,FirstPersonAction_CVPR2018}.
Different human action datasets have also been introduced~\cite{li2010action,wang2012mining,shahroudy2016ntu}.
The datasets differ in the complexity of gestures or actions, the number of individuals, the gesture or action classes, and the types of sensors used for data collection. We selected datasets that provide skeleton data, and we were particularly interested in datasets of small volume. \changetext{This scoping stems from the fact that for large datasets, such as NTU RGB+D 120~\cite{Gupta2021}, data augmentation may not be required.} Therefore, we used the following datasets for evaluation.


\begin{enumerate}
    \item \textbf{MSR Action3D}: The MSR Action3D dataset~\cite{li2010action} provides 20 actions with a total of 567 sequences. 
    We are using the established Protocol A~\cite{nunez2018convolutional} for the training and testing split.
    
    \item \textbf{SHREC'17 Track}: The SHREC'17 Track dataset~\cite{SHREC2017De} has 2,800 sequences which contain 14 gestures performed by 28 individuals \changetext{with one or multiple fingers. The sequences are labeled according to 14 classes if the finger number information is not included or 28 label classes if the finger number information is included (14G and 28G respectively)}. 
    The split between training and testing is defined and well-established in the literature~\cite{SHREC2017De}. 
    \changetext{\item \textbf{DHG-14/28}: The DHG-14/28 dataset~\cite{de2016skeleton} includes 14 gestures with 2,800 sequences provided by 20 individuals. It has the same hand joints, gesture classes and collection method as the SHREC’17 Track dataset. Note that this dataset is only used in the evaluation of model generalizability and impact on state-of-the-art technique performance.
    We use the leave-one-subject-out experimental protocol for training and testing.}
\end{enumerate}




\subsection{Recognition Model}
We need a recognition model to evaluate the augmentation policies. We used two basic well-established effective neural network models, one CNN-based and one LSTM-based, as the recognition models, instead of some hypothetical highly complicated bespoke networks. Our motivation is that this approach serves to demonstrate the overall generalizability of the data augmented from the Imaginative GAN.
To prevent a network bias caused by a specific type of network favoring a specific dataset, we evaluate two different neural networks as the recognition model: LSTM and CNN. 
\changetext{An LSTM is clearly well-suited to modeling the temporal relationships in the sequential skeleton data while a CNN is well-suited to modeling the spatial relationships. To this end, Liu et al.~\cite{liu2020decoupled} explicitly separated the hand gestures into posture variants reflecting their spatial properties and hand movements reflecting their temporal properties. Therefore, these two different networks focus on different aspects of the data and provide breadth to the analysis of the impact of data augmentation.
Each model is configured as follows:}



\begin{enumerate}
    \item \textbf{LSTM}: We adopted the LSTM model by Lai et al.~\cite{lai2018cnn}. We added a self-attention layer before the classification layer. This helps the model determine which part of the network it should give more attention~\cite{cheng2016long}. 
    \item \textbf{CNN}: We adopted the model by Núñez et al.~\cite{nunez2018convolutional} where the structure of the network consists of a CNN attached to a fully-connected
multilayer perceptron. It is a model they use during the pretraining stage. 
\end{enumerate}

Note that both the LSTM and the CNN model are not optimized using techniques such as pretraining CNN models or fine-tuning the hyperparameters (number of hidden units, number of layers, learning rate and dropout rate, etc.). Our focus is on carrying out a fair comparison of the regularization effect brought by the augmented data, not the model.




\subsection{Data Preparation}
The raw data is smoothed using a Savitzky–Golay filter to remove noise. This is achieved by fitting successive subsets of adjacent data points with a low degree polynomial using linear least squares. The window length is set to 7, and the order of the polynomial used to fit the data is set to 3. The smoothed data is then padded with the last row of the data to produce the clean data (CD). Subsequent data augmentations are performed on this clean data.

\subsection{Training}
\subsubsection{Recognition Models}
The models are trained on sparse categorical cross-entropy loss. 
We use Adam as the optimizer and the learning rate is set to 0.0001. The learning rate scheduler is set to be reduced on plateau, and we set the patience to 3. Early stopping is used, and we set the patience to 5. Patience is the number of epochs to wait before reducing the learning rate, or an early stop when no progress is made on the validation set. We use Sparse Categorical Accuracy to evaluate the performance of the model. The training batch size is 64.

\subsubsection{Imaginative GAN} We use the training scheme from the literature~\cite{zhu2017unpaired}. The training batch size is 64. $\lambda_{1}$ and $\lambda_{2}$ in the objective function for the generators are set to 10 and 5 respectively. The number of hidden units for the GRU cell in the generator is 512. We used a computer that has 3 $\times$ Nvidia GeForce RTX 3090 GPU, 1 $\times$ Ryzen 9 3970x CPU.

\begin{figure*}[t]
  \centering
\subfloat[]{\includegraphics[width=0.3\textwidth]{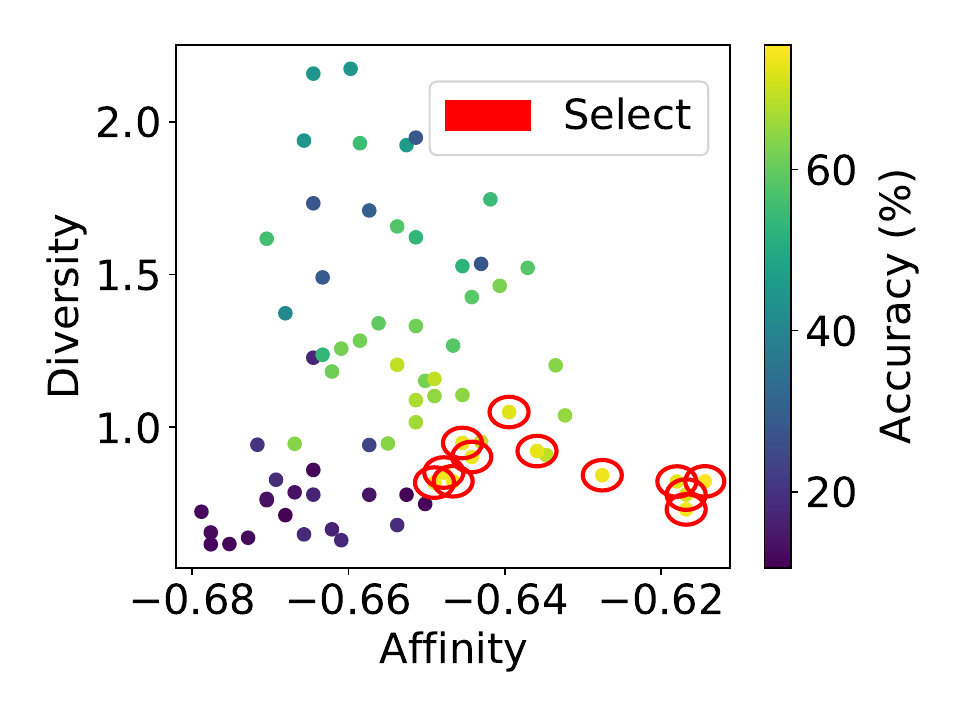}\label{fig:FirstSearch}}
\hfill
\subfloat[]{\includegraphics[width=0.3\textwidth]{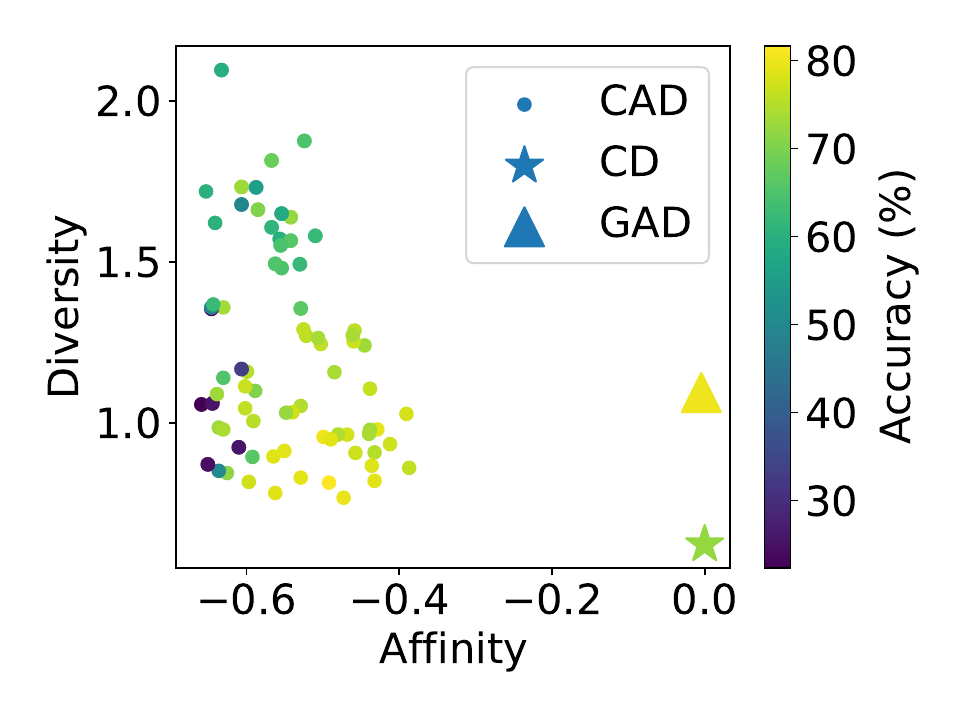}\label{fig:SecondSearch}}
\hfill
\subfloat[ ]{\includegraphics[width=0.3\textwidth]{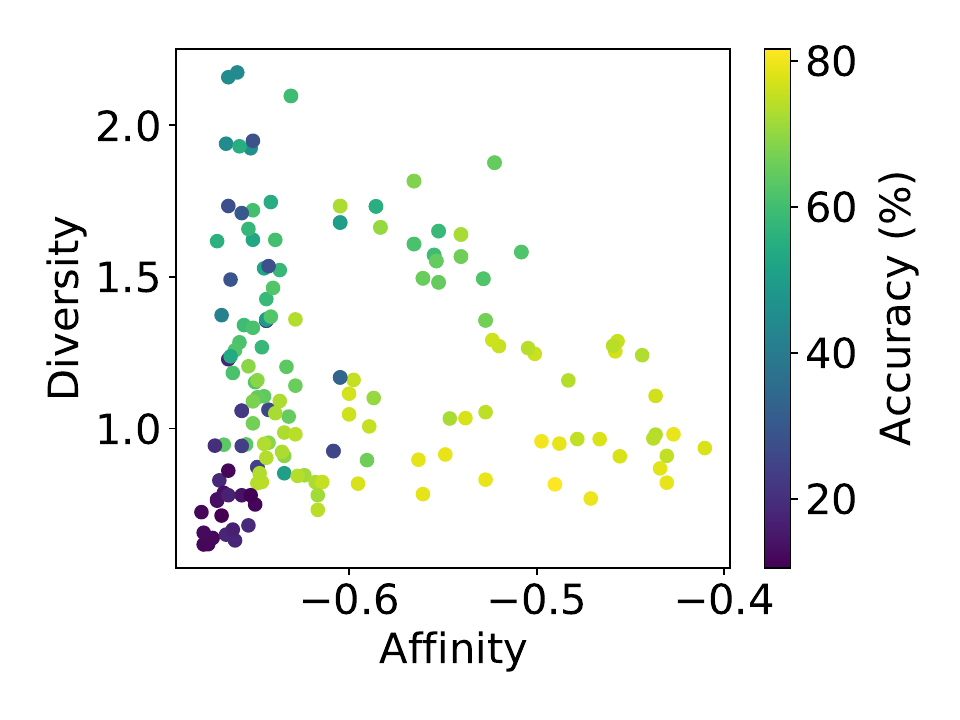}\label{fig:affinity_diversity_cda}}
  \caption{
  (a) The results of the first round search (coarse grid search) for optimal hyperparameters of the classical approach. Circled data points in red are what we will select to inspect and use for a second round search. (b) The results of the second round search (fine grid search). The results from GAD and CD are also shown to compare affinity, diversity, and accuracy. (c) The combination of the results of the first and second round search are presented to show an overview of how affinity and diversity relates to accuracy. 
  We observe that we want affinity to be as high as possible and a diversity at around 1. Results are from experiments when using LSTM as the recognition model and SHREC'17 Track as the evaluation dataset.}
\end{figure*}

\begin{figure}[t]
  \centering
  \includegraphics[width=0.95\linewidth]{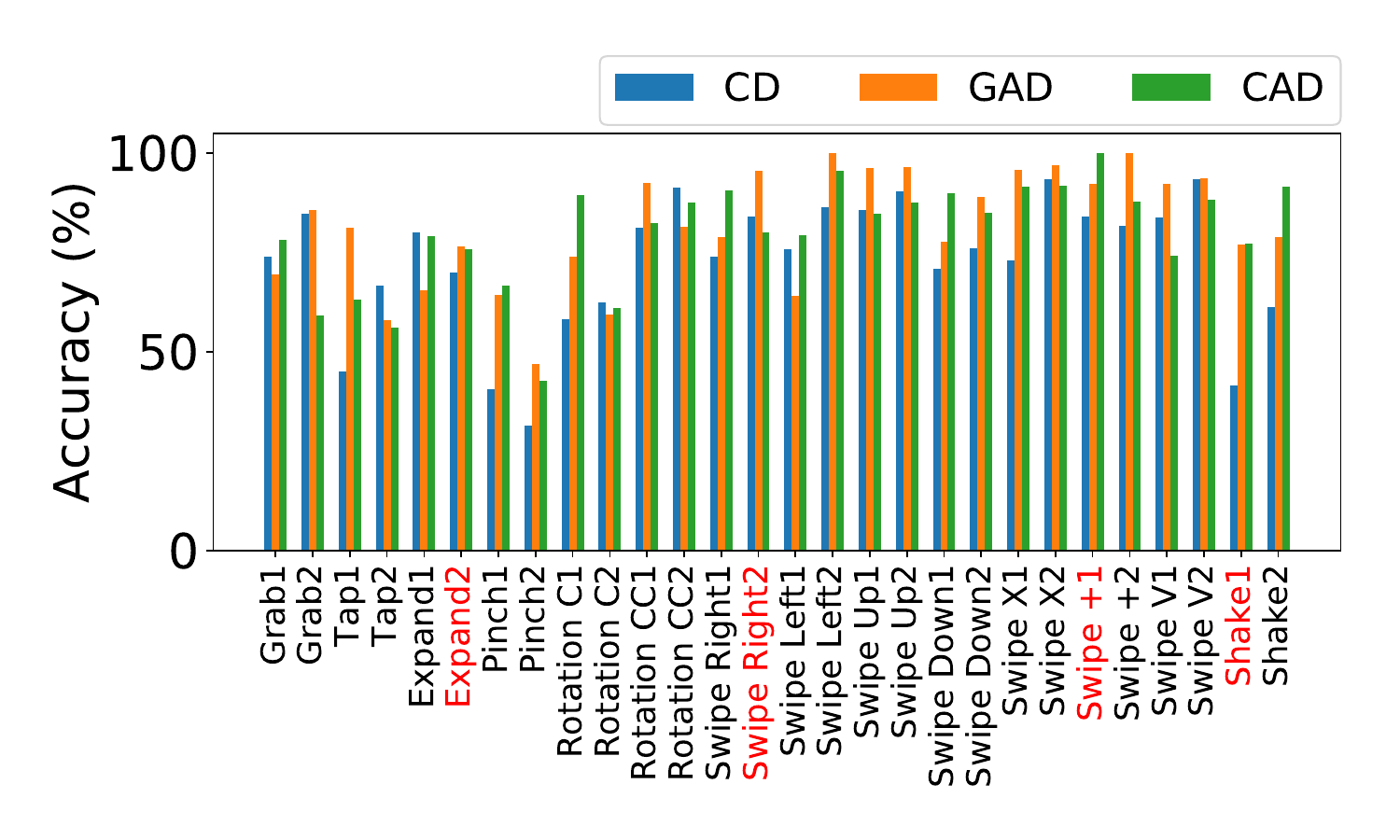}
  \caption{Individual accuracy from the LSTM recognition model of the gestures for the three types of data: clean data (CD), GAN-augmented Data (GAD), and classical augmented data (CAD). Gesture classes in red are the new classes that are excluded from the training of the Imaginative GAN. Synthetic data of the new classes achieved comparable accuracy with the synthetic data of the classes seen in the training phase. The 1 or 2 after the gesture classes indicates the number of fingers used to perform the gesture. 1 represents one finger and 2 indicates the entire hand.}
  \label{fig:accuracy_gesture}
\end{figure}

\section{Results}
\label{result}
Table \ref{table:1} shows the overall results. We can see from the contribution column that the Imaginative GAN is not only fast to train but also leads to augmented data with higher accuracy and more stable performance when trained on the recognition models.
In the remainder of this section, we focus primarily on the results obtained with the LSTM recognition model on the SHREC'17 Track dataset to illustrate the benefits afforded by the Imaginative GAN.


\subsection{Generalization}
\label{generalize}
For the SHREC'17 Track dataset and the LSTM network, we randomly selected four gestures and excluded them in the training of the Imaginative GAN. We then used the generators in the trained Imaginative GAN to generate synthetic data of the four gestures from the real gesture data. Figure \ref{fig:accuracy_gesture} shows that the four gestures on average do not show degradation in their accuracy compared to the clean data and the optimized classical augmentation data. The optimized classical augmentation strategy is discussed in \ref{costtime}. This demonstrates that the model can generalize to data with new classes. We performed the same test with the CNN and MSR Action 3D dataset and also observed this generalization capability.

\changetext{The above result shows that the Imaginative GAN can generalize to new classes. We also performed an evaluation to determine whether the Imaginative GAN can generalize to a new dataset.
To this end we trained the Imaginative GAN on the DHG-14/28 dataset and used this model to generate augmented data for training a CNN and LSTM recognition model for the SHREC'17 dataset. 
The CNN recognition model accuracy was 1.6\% higher than that achieved with the default CAD at 79.6\% while the LSTM model accuracy was 3.2\% higher than the default CAD at 79.8\%.
For both models, the accuracy achieved by the repurposed Imaginative GAN was only just short of that obtained by the GAD when trained on the SHREC'17 dataset.
}

\subsection{Time Efficiency}
\label{costtime}
This investigation also used the SHREC'17 Track dataset and the LSTM network.
In the classical data augmentation approach, there are three hyperparameters to optimize: $\sigma_{scale}$, $\sigma_{shift}$ and $\sigma_{noise}$. There could be more hyperparameters, and this depends on how many transformations are applied to the data. 
Three common ways of performing hyperparameter optimization are random search, grid search, and Bayesian optimization. 
Here we are using grid search to optimize hyperparameters since it is a simple but effective method for exploring a regular search space. We carried out two rounds of search. The first round is a coarse grid search.
We used $\sigma_{scale} \in [0.1,0.15,0.2,0.25,0.3]$, $\sigma_{shift} \in  [0.1,0.15,0.2,0.25,0.3]$ and $\sigma_{noise} \in  [0.1,0.2,0.3]$.

Figure \ref{fig:FirstSearch} shows the result from the first-round search. We observe that the combinations in the red selected circles have good accuracy. We investigated the combinations that produced these results and performed a second round fine grid search, that has $\sigma_{scale} \in [0.1,0.12,0.14,0.18,0.2] $,  $\sigma_{shift} \in[0.1,0.12,0.14,0.18,0.2]$ and $\sigma_{noise} \in [0.05,0.1,0.15]$. 
Figure \ref{fig:SecondSearch} shows the results from the fine grid search together with the validation accuracy achieved from the clean data and GAN augmented data. 
The optimized hyperparameters achieved a validation accuracy of $76.6\%$, which is less than the accuracy ($80.1\%$) achieved by the GAN-augmented data. Compared with the 5 GPU hours training time for the Imaginative GAN, the classical data augmentation approach costs around 15 times more GPU hours (75.2).

\subsection{Escaping Local Minimum}
We now use the CNN model to evaluate the recognition performance of the data (Table \ref{table:1}). We observe that for MSR Action3D, CNN failed to learn the features from CD and CAD, which suggests that the model is trapped in a local optima. In contrast, the GAN-augmented data helped the recognition model to escape the local minimum.




\subsection{Affinity and Diversity}
\label{micrometrics}
\begin{figure}[t]
  \centering
  \subfloat[SHREC'17 Track: Swipe Up with one finger]{\includegraphics[width=0.5\textwidth]{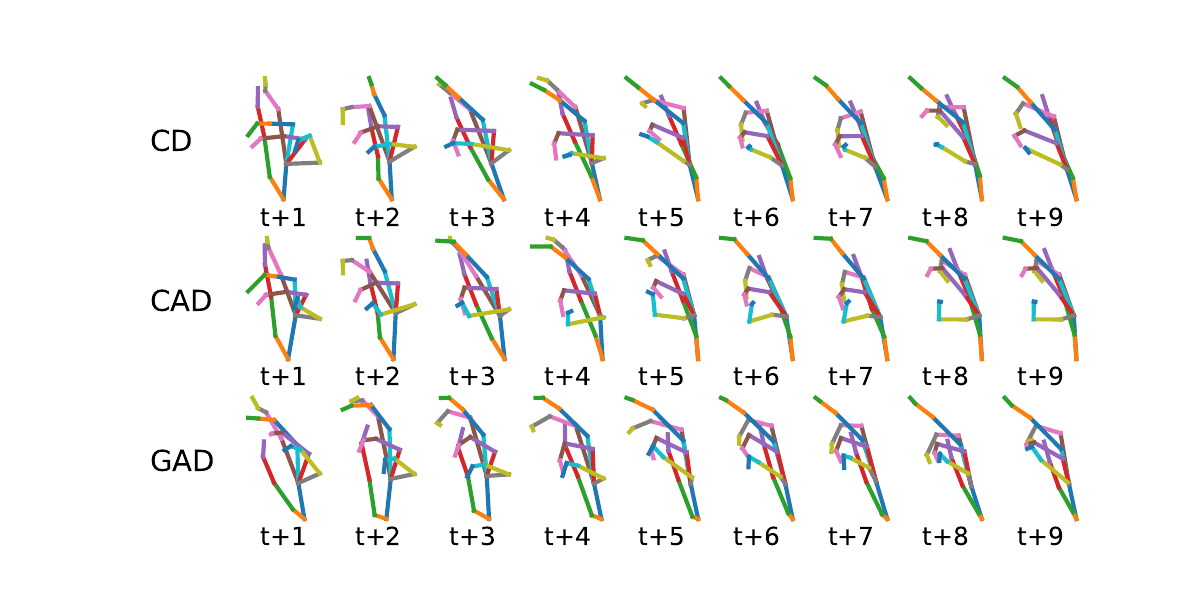}\label{fig:shrec}}
  \hfill
  \subfloat[MSR Action3D: Tennis Swing]{\includegraphics[width=0.5\textwidth]{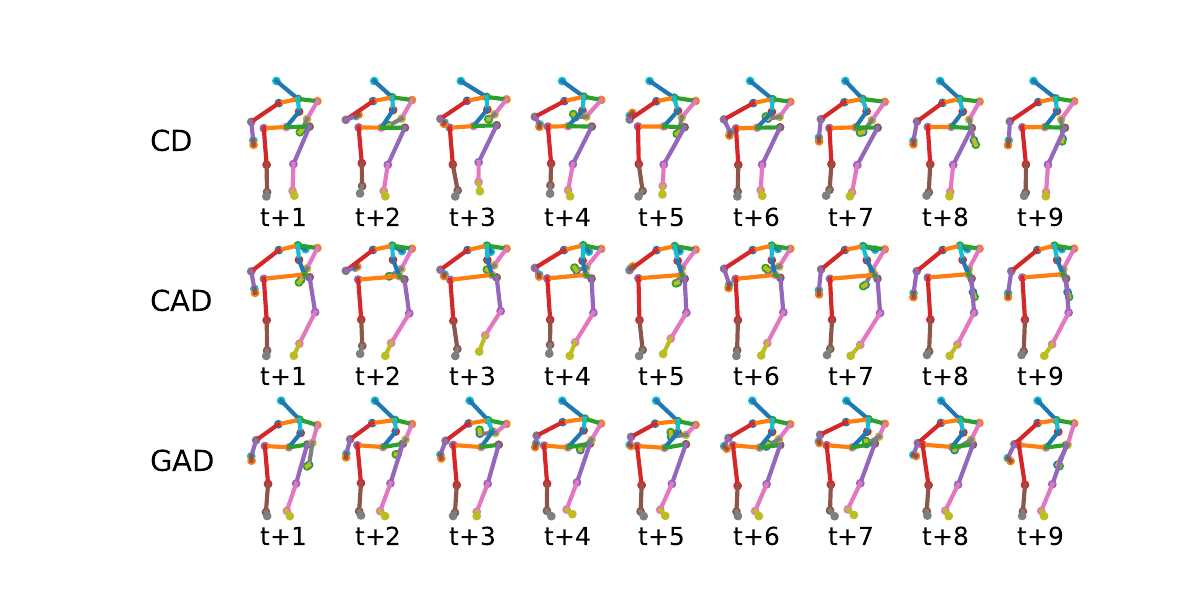}\label{fig:msr}}
  \caption{\changetext{Visualized gesture and action examples illustrating the difference between the clean data and the classical and GAN augmented data. Each timestep represents an even split of the full sequence.}}
\label{fig:visual}
\end{figure}

For the SHREC'17 Track dataset and the LSTM network, in Figure \ref{fig:affinity_diversity_cda}, we observe, as expected, that the points with high affinity and diversity around 1.0 have the highest accuracy, and data with low affinity and low diversity cannot produce a good result. We also observe that a diversity that is too high can decrease accuracy as the model may fail to learn the complexity of the data. Therefore, we expect that data with as high affinity as possible and diversity of around 1.0 will achieve the best recognition accuracy. This is achieved by the GAN-augmented data. The results from the CNN and the MSR Action3D dataset were found to highlight the same advantages of GAD over CD and CAD. \changetext{Both GAD and CAD are visualized in Figure \ref{fig:visual} in their optimized settings. We see that GAD maintains a realistic spatial structure while CAD produces some unrealistic temporal and spatial alterations. 
Further, GAD achieves not only good geometric diversities but also balanced temporal correlations.}

\subsection{Ablation Study}
We performed an ablation study on the MSR Action3D dataset to evaluate the importance of the number of hidden units of the GRU cell in the Imaginative GAN. We tested four numbers of hidden units: 64, 128, 256 and 512. They lead to a validation accuracy of $46.4\%$, $55.1\%$, $64.2\%$ and $67.3\%$ respectively. This result can be justified by the reasoning that more hidden units can encode more information of the input sequence. A higher number of hidden units only increases the training time by within 10\% which is acceptable. Therefore, 512 hidden units are used for a higher validation accuracy. 


\subsection{Impact on State-of-the-Art Recognition Methods}
\changetext{
The results presented in the previous subsections illustrate the benefits of the Imaginative GAN and GAN augmented data on two basic but well-established recognition models (CNN and LSTM). Here we show that GAN augmented data can also benefit state-of-the-art methods for skeleton-based data discrimination.
For this purpose, we performed a search for state-of-the-art methods for skeleton-based gesture recognition, with the additional criteria that the actual recognition performance is reported and that the code is publicly available. 
This search yielded the following methods: i) Dynamic Graph-Based Spatial-Temporal Attention (DG-STA)~\cite{chen2019construct}; ii) Double-Feature Double-Motion Network (DD-Net)~\cite{yang2019make}; and iii) Decoupled Spatial-Temporal Attention Network (DSTA-Net)~\cite{shi2020decoupled}.
Table~\ref{table:sota} summarizes the benefits provided by the GAN-based augmentation method in improving upon the recognition accuracy reported in the corresponding papers. 
Note that where any of these methods were already using a data augmentation method, it was removed and replaced by GAD. 
The results shown in Table~\ref{table:sota} highlight the broad utility of the data augmentation approach.} 

\begin{table}[t]
\centering

\begin{tabular}{ccccc}
\hline
\multirow{2}{*}{Model}           & \multicolumn{2}{c}{SHREC'17 Track} & \multicolumn{2}{c}{DHG-14/28} \\ \cline{2-5} 
                                 & 14G              & 28G             & 14G           & 28G           \\ \hline
DG-STA~\cite{chen2019construct}  & +0.7\%           & +0.8\%          & +0.6\%        & +0.4\%        \\
DD-Net~\cite{yang2019make}       & +0.7\%           & 0.7\%           & +0.6\%        & +0.6\%        \\
DSTA-Net~\cite{shi2020decoupled} & +1.0\%           & 0.7\%           & +0.8\%        & +0.5\%        \\ \hline
\end{tabular}

\caption{\changetext{Impact of augmented data from the Imaginative GAN on the performance of state-of-the-art methods. 
The values indicate the benefit afforded by GAD over the peak accuracy achieved in the corresponding publication. Note that each datasets has two ways of data labelling according to the finger used (14G and 28G).}} 
\label{table:sota}
\end{table}

\section{Limitations and Future Work}
\label{limitation}
A potential limitation is the requirement for some tuning of the learning rate and number of hidden units in the Imaginative GAN. However, in contrast to the huge hyperparameter space for the classical augmentation approach, tuning just two hyperparameters is comparatively straightforward and fast. 
If a trained model exists, then generating thousands of augmented data samples takes less than a second. 

In theory, the Imaginative GAN is able to approximate any trajectory. Trajectory data records locations of moving objects at certain moments, and has been used widely when researching, for example, human behavior and traffic problems. We believe fruitful future work is to verify this statement by carrying out experiments on generating sequences of behaviors/movement from other domains. 


\section{Conclusions}
\label{conclusion}
We have proposed an automatic data augmentation tool that can `imagine' realistic alterations to input data. 
The data augmented using the Imaginative GAN is evaluated against data augmented using a classical approach as well as the denoised and padded raw data. Our results have demonstrated that the Imaginative GAN outperforms the classical approach in mean validation accuracy, standard error, and in the time cost to find an optimized strategy.


In summary, the Imaginative GAN is superior to simply adding stochastic and sequential distortions to data. In the future, we foresee a library of Imaginative GANs trained on different types of data, allowing researchers to directly download pre-trained Imaginative GANs and instantly generating their own synthetic data. While the classical approach may still be a useful tool to prevent over-fitting and increase the deep learning model's performance, the Imaginative GAN can always be used as a baseline to compare against, since it takes almost zero effort and zero prior knowledge or inspection of the input data to generate synthetic data using the Imaginative GAN. 






{\small
\bibliographystyle{ieee}
\bibliography{egbib}
}

\end{document}